\pdfoutput=1

\documentclass[11pt]{article}

\usepackage[final]{acl}

\usepackage{times}
\usepackage{latexsym}

\usepackage[T1]{fontenc}

\usepackage[utf8]{inputenc}

\usepackage{microtype}

\usepackage{inconsolata}

\usepackage{graphicx}

\usepackage{amsmath, amsfonts, amsthm, bm}

\def\eqref#1{equation~\ref{#1}}

\def\1{\bm{1}}

\DeclareMathAlphabet{\mathsfit}{\encodingdefault}{\sfdefault}{m}{sl}
\SetMathAlphabet{\mathsfit}{bold}{\encodingdefault}{\sfdefault}{bx}{n}

\def\BR{\textsc{br}}
\def\GP{\textsc{gp}}
\def\CR{\textsc{cr}}
\def\LR{\textsc{lr}}

\definecolor{layercolor}{HTML}{E5C100}
\definecolor{brcolor}{HTML}{DA9876}
\definecolor{gpcolor}{HTML}{4E95D9}
\definecolor{crcolor}{HTML}{DB4447}

\usepackage{enumitem}
\usepackage{amsmath}
\usepackage{amssymb}
\usepackage{mathtools}
\usepackage{amsthm}
\usepackage{mathabx}

\usepackage[export]{adjustbox}

\usepackage[capitalize]{cleveref}
\Crefname{section}{Section}{Sections}
\Crefname{table}{Table}{Tables}
\Crefname{figure}{Figure}{Figures}

\makeatletter
\def\@footnotecolor{red}
\define@key{Hyp}{footnotecolor}{%
 \HyColor@HyperrefColor{#1}\@footnotecolor%
}
\def\@footnotemark{%
    \leavevmode
    \ifhmode\edef\@x@sf{\the\spacefactor}\nobreak\fi
    \stepcounter{Hfootnote}%
    \global\let\Hy@saved@currentHref\@currentHref
    \hyper@makecurrent{Hfootnote}%
    \global\let\Hy@footnote@currentHref\@currentHref
    \global\let\@currentHref\Hy@saved@currentHref
    \hyper@linkstart{footnote}{\Hy@footnote@currentHref}%
    \@makefnmark
    \hyper@linkend
    \ifhmode\spacefactor\@x@sf\fi
    \relax
  }%
\makeatother

\hypersetup{
    colorlinks = true,
    linkcolor = red,
    anchorcolor = red,
    urlcolor = red,
    citecolor = blue
}

\usepackage{subcaption}     
\usepackage{multirow,multicol}
\usepackage{booktabs}

\usepackage{xcolor}
\usepackage{xspace}
\makeatletter
\DeclareRobustCommand\onedot{\futurelet\@let@token\@onedot}
\def\@onedot{\ifx\@let@token.\else.\null\fi\xspace}
\def\eg{\emph{e.g}\onedot} 
\def\ie{\emph{i.e}\onedot} \def\Ie{\emph{I.e}\onedot}

\makeatother
\usepackage{wrapfig}

\title{Rethinking Pruning Large Language Models:\\Benefits and Pitfalls of Reconstruction Error Minimization}

\makeatletter
\renewcommand{\@fnsymbol}[1]{\textcolor{black}{\ifcase#1\or *\or \dagger\or \ddagger\or
\mathsection\or \mathparagraph\or \|\or **\or \dagger\dagger
\or \ddagger\ddagger \fi}}
\makeatother

\author{Sungbin Shin$^1$\thanks{Work partly done as a student researcher at Google} \quad Wonpyo Park$^2$ \quad Jaeho Lee$^{1,2,3}$ \quad Namhoon Lee$^{1,2,3}$\\
$^1$POSTECH \quad $^2$Google \quad $^3$Yonsei University\\
\texttt{\{ssbin4,jaeho.lee,namhoonlee\}@postech.ac.kr} \\
\texttt{wppark@google.com}
}

\begin{document}
\maketitle

\begin{abstract}

This work suggests fundamentally rethinking the current practice of pruning large language models (LLMs).
The way it is done is by divide and conquer: split the model into submodels, sequentially prune them, and reconstruct predictions of the dense counterparts on small calibration data one at a time; the final model is obtained simply by putting the resulting sparse submodels together.
While this approach enables pruning under memory constraints, it generates high reconstruction errors.
In this work, we first present an array of reconstruction techniques that can significantly reduce this error by more than $90\%$.
Unwittingly, however, we discover that minimizing reconstruction error is not always ideal and can overfit the given calibration data, resulting in rather increased language perplexity and poor performance at downstream tasks.
We find out that a strategy of self-generating calibration data can mitigate this trade-off between reconstruction and generalization, suggesting new directions in the presence of both benefits and pitfalls of reconstruction for pruning LLMs.\footnote{Our code is available at \url{https://github.com/LOG-postech/rethinking-LLM-pruning}.}

\end{abstract}
\section{Overview}

Large language models (LLMs) have shown remarkable potential and achieved tremendous successes in various domains \citep{brown2020language,singhal2023large,roziere2023code}.
Nevertheless, running them requires a significant amount of computations and memory, raising concerns about accessibility, sustainability, and scalability \citep{strubell2019energy,bender2021dangers}.
Neural network pruning holds great promise for mitigating this issue \citep{lecun1989optimal,hoefler2021sparsity}.
A complication here is that the standard approach is not quite feasible since it usually involves an extensive training process (and training data) which is challenging to carry out for LLMs.

\begin{figure}[!t]
    \centering
    \begin{subfigure}{\linewidth}
    \vspace{0.5em}
1        \includegraphics[width=\linewidth]{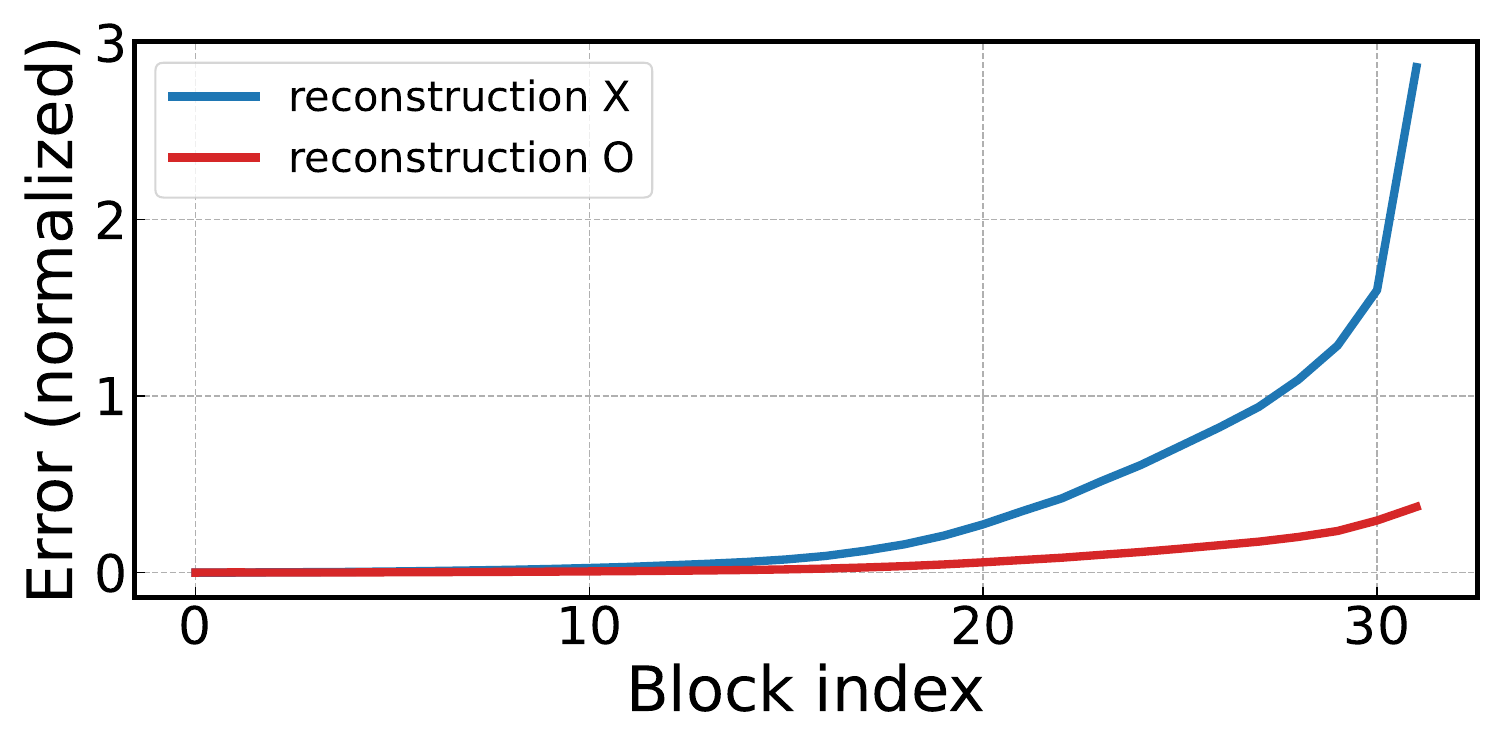}
        \caption{Effects of reconstruction techniques on reducing the error}
        \label{fig:intro-potential}
    \end{subfigure}
    \begin{subfigure}{\linewidth}
        \vspace{0.5em}
        \includegraphics[width=\linewidth,trim={8em 0 13em 0},clip]{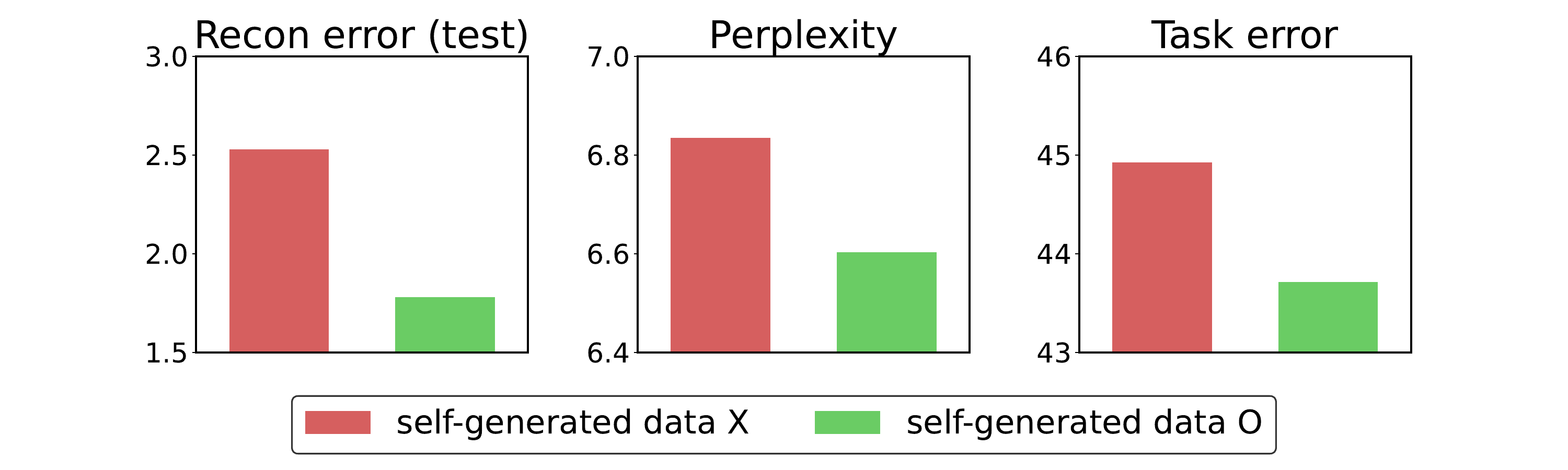}
        \caption{Effects of self-generated data on mitigating overfitting}
        \label{fig:intro-pitfall}
    \end{subfigure}
    \caption{
    (a)
    Reconstruction techniques significantly reduce the compounding errors and lead to a substantial reduction of error in the final block.
    Reconstruction \textsc{o} and \textsc{x} refer to the results with and without the proposed reconstruction techniques (\BR{}, \GP{}, \CR{}) respectively.
    (b)
    Minimizing reconstruction error may not always be ideal since models can overfit calibration data (we show this in \cref{sec:exp-generalization}).
    Using our self-generated calibration data in the reconstruction process mitigates this issue quite effectively by decreasing test error, perplexity, and error rates for downstream tasks.
    }
    \label{fig:intro-recon-error}
\end{figure}

To address this issue, LLM pruning is done post training.
Specifically, it could be formulated as a reconstruction problem as follows:
\begin{equation}
\label{eq:reconstruction}
\begin{aligned}
& \underset{w,m}{\mathrm{min}} & & \left\|f(\bar{w}; \mathcal{D}) - f(m \odot w; \mathcal{D})\right\|_2^2 \\
& \mathrm{s.t.} & & \|m\|_0 \le k \ ,
\end{aligned}
\end{equation}
\ie, given a pre-trained model $\bar{w}$, the goal is to find a pruning mask $m$ such that the resulting sparse model $m \odot w$ reconstructs the predictions of the original dense model $f(\bar{w};\cdot)$ on some calibration data $\mathcal{D}$;
here, $\odot$ denotes element-wise product for vectorized representations, and $m$ needs to satisfy a given sparsity constraint $k$.
If the objective criterion---\emph{reconstruction error}---is minimized to zero, then we achieve the perfect reconstruction and thereby pruning results.

\begin{figure*}[!th]
    \centering
    \includegraphics[width=0.7\linewidth,trim={0 7em 0 13em},clip]{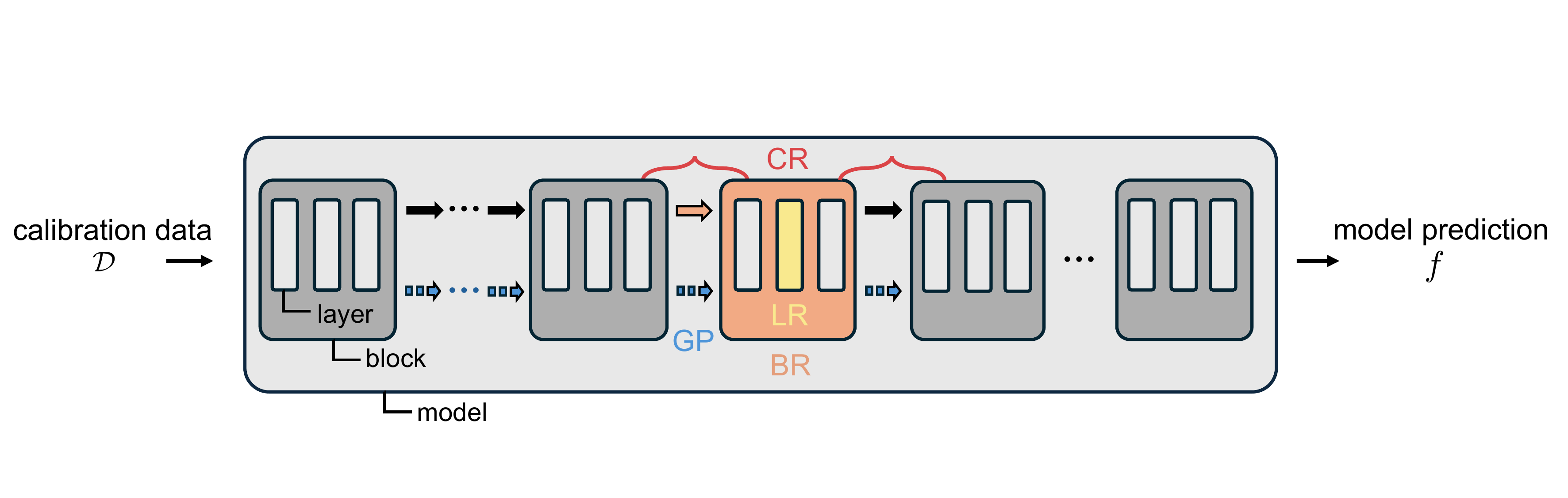}
    \caption{
    An illustration of reconstruction techniques for pruning large language models.
    Here, we want the sparse model $f(m \odot w; \cdot)$ to reconstruct the prediction of the dense model on some calibration data $\mathcal{D}$.
    \textcolor{layercolor}{\textsc{lr}}, \textcolor{brcolor}{\BR{}}, \textcolor{gpcolor}{\GP{}}, and  \textcolor{crcolor}{\CR{}} each correspond to layer-wise reconstruction, block-wise reconstruction, global propagation, and cross-block reconstruction.
    Here, solid and dashed arrows each represent the inputs coming from sparse and dense models.
    }
    \label{fig:techniques}
\end{figure*}

While one could now avoid training LLMs from scratch with (\ref{eq:reconstruction}), it still requires as much memory as of the given LLM, hindering development under memory constraints.
To circumvent this issue, many recent works take a divide-and-conquer approach:
\ie, split the model into a sequence of smaller submodels, prune and reconstruct each submodel individually, and simply put all resulting sparse submodels together \citep{frantar2023sparsegpt,sun2023simple,zhang2023dynamic}.
Albeit fairly effective, we find that this can easily create critically high compounding errors.
This is because solutions for each subproblem yield non-zero reconstruction errors.

In this work, we address the reconstruction error minimization for pruning LLMs with the following three major pillars.
First, we focus on developing various engineering techniques to reduce this error.
These are inspired to lessen the suboptimality of subsolutions by incorporating different levels of extension schemes.
Second, we suggest that reducing this error is not necessarily favorable, however.
Our extensive experimental results indicate that it is possibly due to overfitting, given limited calibration data and high problem complexity.
Third, we present useful strategies to potentially mitigate the risk of reconstruction and improve generalization.
This is based on what we call the self-generation of calibration data.

Briefly, this work investigates the benefits and pitfalls of the reconstruction error minimization scheme for pruning LLMs.
To our best knowledge, this trade-off has not been explicitly identified or studied before, thereby suggesting rethinking the current practice.
Our initial investigations may shed light on some potential future research directions.
We summarize our main results in \cref{fig:intro-recon-error}.
\section{Reconstruction Techniques}
\label{sec:reconstruction-methods}

This section explains three optimization schemes we use to reduce reconstruction errors in this work.

\paragraph{Block-wise reconstruction (\BR{})}

The seminal work of \citet{frantar2023sparsegpt} proposes to reconstruct predictions layer-wise based on least squares.
By removing non-linearity this approach yields a closed-form solution.
However, we find that this can create a high reconstruction error since the system is highly underdetermined (\ie, there are much more parameters than calibration data).
To reduce compounding errors, we first consider extending the unit of optimization target from a layer to a block of layers.
Specifically, this means a block-wise reconstruction (\BR{}) which can be formulated as follows:
\begin{equation}
\label{eq:reconstruction-block}
\underset{w_1,\hdots,w_B}{\mathrm{min}} \sum_{i=1}^{B} \left\|g_i (\bar{w}_i; x_i) - g_i (\bar{m}_i \odot w_i; x_i)\right\|_2^2
\end{equation}
where $g_i$ refers to the $i$-th block of layers (\eg, a Transformer block) in which we have the optimization variables $w_i$, and $x_i$ denotes the inputs to the $i$-th block which originally come from calibration data;
here, the pruning mask $\bar{m}$ is fixed assuming that it is already obtained from an arbitrary pruning method.
\Ie, the goal is to update variables in each block to minimize the extended reconstruction errors.
We solve this problem iteratively using the standard gradient-based method.
Notably a similar approach is also proposed in the concurrent work of \citet{guo2024ebft}, and we find in our experiments that \BR{} is extremely effective in reducing the reconstruction errors in \cref{sec:exp-reconerror}.
We illustrate the idea of \BR{} in \cref{fig:techniques}.

\paragraph{Global propagation (\GP{})}

While the general divide-and-conquer principle is quite functional, we identify a potential issue therein: by sequentially solving the subproblem, it is constantly fitting practically suboptimal solutions obtained from the previous step (which become gradually worse), as with $x_i = g_{i-1} (\bar{m}_{i-1} \odot w_{i-1}; x_{i-1})$.
We realize that this is another source of compounding errors, and thus, suggest that when we locally reconstruct a model, at least we use global propagation (\GP{}) from the original dense model as input to the target reconstruction; \ie, $x_i = g_{i-1}(\bar{w}_{i-1}; x_{i-1})$.
We show that \GP{} improves the reconstruction results quite significantly in \cref{sec:exp-reconerror}.
We further note that a similar principle is found in various applications including low-rank approximation \citep{zhang2015accelerating}, channel pruning \citep{he2017channel}, and quantization \citep{nagel2020up,hubara2021accurate}.
We illustrate the idea of \GP{} in \cref{fig:techniques}.

\paragraph{Cross-block reconstruction (\CR{})}

Another way we consider to further reduce reconstruction errors is to extend the reconstruction unit from a block to multiple blocks and stitch the solutions in between by connecting via the adjacent blocks.
Specifically, this means that now $g$ in (\ref{eq:reconstruction-block}) becomes a composite of multiple blocks, say $h$, and we ensure $h$ overlaps;
more precisely, $h_i = g_{i} \circ g_{i-1}$ and $h_{i+1} = g_{i+1} \circ g_{i}$ for two blocks, and so on for all blocks.
This way, namely cross-block reconstruction or \CR{} \citep{ding2023cbq}, we can potentially bridge between subsolutions by taking into account some interaction between adjacent blocks, and hence, reduce the compounding errors.
We illustrate the idea of \CR{} in \cref{fig:techniques}.

To elaborate further, the difference between \BR{} and \CR{} is that while \BR{} is about updating parameters within a block (thus it is not concerned with how to combine subsolutions), \CR{} takes a step further and is about stitching the subsolutions;
\ie, \CR{} updates parameters within two adjacent blocks, and when it comes to reconstructing the next block, it includes the overlapping block so that it has the effect of ``stitching''.
This method is found to be quite effective for reducing the error, however, we find that this method can often lead to overfitting.
We discuss this in detail in \cref{sec:exp-generalization}.
\section{Experiments}
\label{sec:experiments}

\subsection{Reconstruction error}
\label{sec:exp-reconerror}

\begin{figure}[!t]
    \centering
    \begin{subfigure}[b]{0.475\linewidth}
        \includegraphics[width=\linewidth]{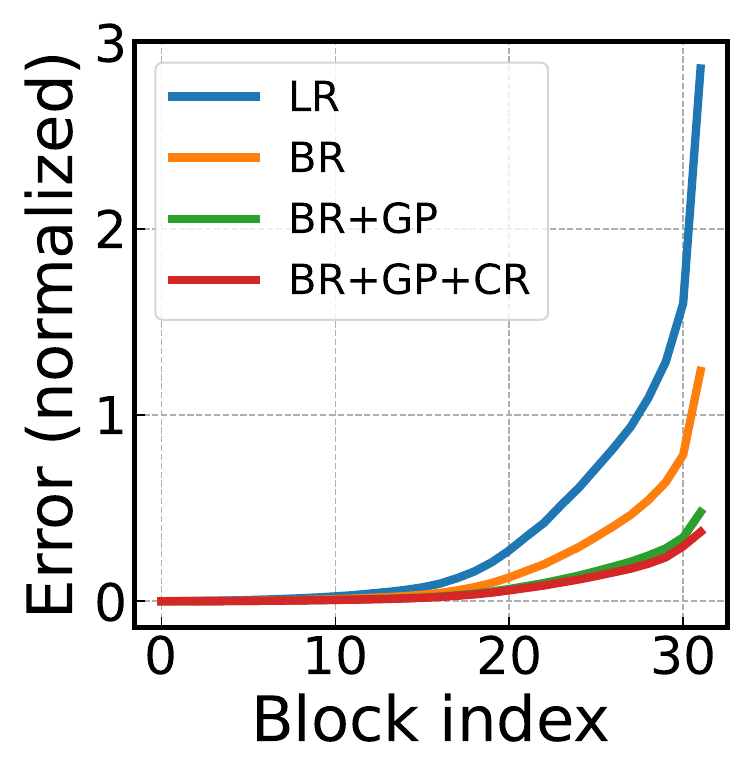}
        \caption{SparseGPT}
         \label{fig:recon-error-sparsegpt}
    \end{subfigure}
    \begin{subfigure}[b]{0.48\linewidth}
        \includegraphics[width=\linewidth,trim={0 0.1em 0 0},clip]{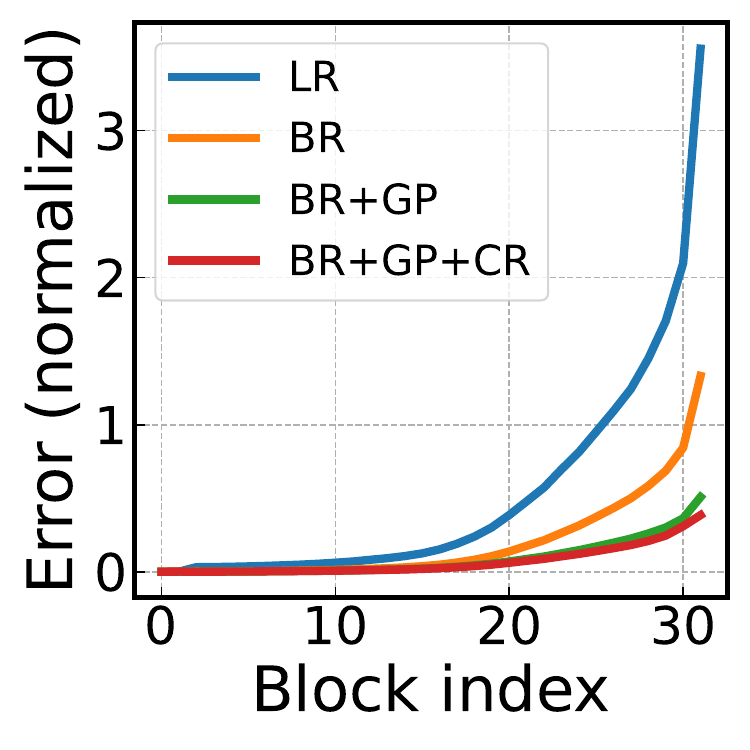}
        \caption{Wanda}
         \label{fig:recon-error-wanda}
    \end{subfigure}
    \caption{
    Results of reconstruction techniques for LLaMA-7B.
    They constantly reduce the compounding errors, achieving a significant decrease at the final block ($\sim 90\%$).
    We find this trend is consistent across different settings.
    See \cref{fig:recon-error-app,fig:recon-error-app-opt} of \cref{sec:app-results} for more results.
    }
    \label{fig:recon-error-exp}
\end{figure}

We first evaluate the effectiveness of the suggested techniques in reducing the reconstruction error.
Here, we focus on pruning LLaMA-7B \citep{touvron2023llama} and OPT-125M \citep{zhang2022opt} to unstructured $50\%$ sparsity with three pruning methods: SparseGPT \citep{frantar2023sparsegpt}, Wanda \citep{sun2023simple}, and Magnitude \citep{han2015learning}.
For each pruning method, we examine four reconstruction strategies: layer-wise reconstruction (\LR{}), block-wise reconstruction (\BR{}), block-wise reconstruction with global propagation (\BR{}+\GP{}), and cross-block reconstruction with global propagation (\BR{}+\GP{}+\CR{}).
Following the convention, we use $256$ calibration data randomly sampled from C4 \citep{raffel2020exploring} each containing $1024$ tokens.
We run the Adam optimizer for $10$ epochs (see \cref{sec:app-details} for details).
The results are presented in \cref{fig:recon-error-exp}.

We can see that all the reconstruction techniques reduce the compounding errors quite significantly, yielding a substantial reduction at the final block.
Specifically, \BR{} first reduces the final error by at least $50\%$ across all pruning methods compared to \LR{}, \BR{}+\GP{} further reduces the error by at least $60\%$ compared to \BR{}, and finally, \BR{}+\GP{}+\CR{} reduces the error by at least $20\%$ compared to \BR{}+\GP{}.
Consequently, we observe that the error is reduced from $87\%$ to $94\%$ with \BR{}+\GP{}+\CR{} compared to the baseline \LR{}.

\subsection{Generalization performance}

\label{sec:exp-generalization}

\begin{table*}[!t]
    \centering
    \resizebox{\linewidth}{!}{
    \begin{tabular}{c c | c | c c c c| c c c c c c c c}
      \toprule
      \multirow{2}{*}{Pruner} & \multirow{2}{*}{Reconstruction} & \multirow{2}{*}{Error (normalized)} & \multicolumn{4}{c|}{Perplexity} & \multicolumn{8}{c}{Zero-shot accuracy} \\
      & & & Wiki & PTB & C4 & Mean & BoolQ & RTE & HellaSwag & WinoGrande & ARC-e & ARC-c & OpenbookQA & Mean \\
      \midrule
      Dense & $-$ & $-$ & $5.68$ & $10.12$ & $7.34$ & $7.71$ & $75.11$ & $66.43$ & $56.96$ & $70.00$ & $75.29$ & $41.81$ & $34.40$ & $60.00$\\
      \midrule
      \multirow{4}{*}{SparseGPT}& \LR{} & $2.86$ & $7.24$ & $12.61$ & $9.17$ & $9.67$ & $\underline{73.36}$ & $\underline{58.12}$ & $51.86$ & $\underline{68.90}$ & $70.62$ & $36.95$ & $\underline{28.60}$ & $\mathbf{55.49}$\\
& \BR{} & $1.24$ & $6.82$ & $11.69$ & $8.66$ & $9.06$ & $71.71$ & $54.51$ & $52.54$ & $68.27$ & $71.68$ & $36.18$ & $28.40$ & $54.76$\\
& \BR{}+\GP{} & $0.48$ & $\underline{6.72}$ & $\underline{11.32}$ & $\underline{8.55}$ & $\mathbf{8.86}$ & $71.22$ & $53.79$ & $\underline{53.57}$ & $68.90$ & $\underline{71.76}$ & $\underline{37.54}$ & $27.80$ & $54.94$\\
& \BR{}+\GP{}+\CR{} & $\mathbf{0.37}$ & $6.83$ & $11.41$ & $8.71$ & $8.99$ & $72.91$ & $55.60$ & $53.24$ & $68.51$ & $71.21$ & $36.26$ & $27.80$ & $55.07$\\
      \midrule
      \multirow{4}{*}{Wanda}&  \LR{} & $3.56$ & $7.25$ & $12.77$ & $9.28$ & $9.77$ & $71.28$ & $55.23$ & $52.04$ & $66.46$ & $69.36$ & $36.52$ & $28.80$ & $54.24$\\
& \BR{} & $1.33$ & $6.82$ & $11.54$ & $8.70$ & $9.02$ & $72.02$ & $57.04$ & $52.45$ & $67.09$ & $\underline{72.18}$ & $36.60$ & $28.60$ & $55.14$\\
& \BR{}+\GP{} & $0.51$ & $\underline{6.68}$ & $\underline{11.25}$ & $\underline{8.56}$ & $\mathbf{8.83}$ & $72.66$ & $\underline{60.29}$ & $\underline{53.25}$ & $\underline{68.43}$ & $71.46$ & $\underline{37.63}$ & $\underline{29.80}$ & $\mathbf{56.22}$\\
& \BR{}+\GP{}+\CR{} & $\mathbf{0.38}$ & $6.79$ & $12.01$ & $8.72$ & $9.18$ & $\underline{73.00}$ & $59.93$ & $53.18$ & $68.27$ & $71.13$ & $37.29$ & $28.80$ & $55.94$\\
      \midrule
      \multirow{4}{*}{Magnitude} & \LR{} & $8.08$ & $17.29$ & $49.67$ & $23.78$ & $30.25$ & $54.65$ & $\underline{54.15}$ & $45.47$ & $59.43$ & $58.75$ & $33.45$ & $22.60$ & $46.93$\\
& \BR{} & $2.37$ & $7.83$ & $15.73$ & $9.66$ & $11.07$ & $68.90$ & $49.82$ & $47.85$ & $66.38$ & $70.29$ & $36.77$ & $27.00$ & $52.43$\\
& \BR{}+\GP{} & $0.63$ & $\underline{6.88}$ & $\underline{11.77}$ & $\underline{8.77}$ & $\mathbf{9.14}$ & $71.65$ & $52.35$ & $53.00$ & $\underline{68.19}$ & $\underline{70.75}$ & $\underline{37.63}$ & $\underline{29.00}$ & $\mathbf{54.65}$\\
& \BR{}+\GP{}+\CR{} & $\mathbf{0.46}$ & $6.98$ & $11.96$ & $8.85$ & $9.27$ & $\underline{72.23}$ & $48.74$ & $\underline{53.20}$ & $67.09$ & $70.54$ & $36.95$ & $28.20$ & $53.85$\\
      \bottomrule
  \end{tabular}
  }
  \caption{
  Effects of different reconstruction techniques on error, perplexity, and zero-shot accuracy for LLaMA-7B.
  \textbf{Bold} and \underline{underline} refer to best in general and task-specific.
  See \cref{tab:performance-opt} of \cref{sec:app-results} for the OPT-125M results.
  }
  \label{tab:performance}
\end{table*}

\begin{table}
   \centering
   \captionsetup[subtable]{position = below}
   \begin{subtable}{0.48\linewidth}
       \centering
       \resizebox{\linewidth}{!}{
    \begin{tabular}{c c c c}
      \toprule
     \multirow{2}{*}{Pruner} & \multirow{2}{*}{\CR{}} & \multicolumn{2}{c}{Error (normalized)} \\
      & & Calib & Test \\
      \midrule
      \multirow{2}{*}{SparseGPT} & X & $0.006$ & $0.0083$\\
& O & $\mathbf{0.004}$ & $\mathbf{0.0078}$\\
      \midrule
      \multirow{2}{*}{Wanda} 
& X & $0.006$ & $0.0080$\\
& O & $\mathbf{0.004}$ & $\mathbf{0.0076}$\\
      \midrule
      \multirow{2}{*}{Magnitude} 
& X & $0.008$ & $0.0109$\\
& O & $\mathbf{0.005}$ & $\mathbf{0.0102}$\\
      \bottomrule
  \end{tabular}
  }
   \caption{OPT-125M}
   \label{tab:testerror-opt}
   \end{subtable}%
  \hspace{0.2em}
   \begin{subtable}{0.48\linewidth}
       \centering
       \resizebox{\linewidth}{!}{
    \begin{tabular}{c c c c}
      \toprule
      \multirow{2}{*}{Pruner} & \multirow{2}{*}{\CR{}} & \multicolumn{2}{c}{Error (normalized)} \\
      & & Calib & Test \\
      \midrule
      \multirow{2}{*}{SparseGPT} & X & $0.48$ & $\mathbf{2.30}$\\
& O & $\mathbf{0.37}$ & $2.53$\\
      \midrule
      \multirow{2}{*}{Wanda} 
& X & $0.51$ & $\mathbf{2.23}$\\
& O & $\mathbf{0.38}$ & $2.48$\\
      \midrule
      \multirow{2}{*}{Magnitude} 
& X & $0.63$ & $\mathbf{2.42}$\\
& O & $\mathbf{0.46}$ & $2.55$\\
      \bottomrule
  \end{tabular}
  }
   \caption{LLaMA-7B}
   \label{tab:testerror-llama7B}
   \end{subtable}%
   \caption{
   Reconstruction errors of OPT-125M and LLaMA-7B on test data (raw-Wikitext2) as well as calibration data.
   Overfitting by \CR{} is only observed for the larger LLaMA-7B model.
   We find that larger models in general are more susceptible to overfitting.
   See \cref{tab:performance-opt,tab:testerror-app} of \cref{sec:app-results} for more results.
   }
   \label{tab:testerror}
\end{table}

We now evaluate the generalization performances of the reconstruction results.
Specifically, we measure the perplexity of the pruned model on three different datasets: raw-Wikitext2 \citep{merity2016pointer}, PTB \citep{marcus1994penn}, and validation data of C4.
We also measure its zero-shot task performance in accuracy on seven downstream tasks: BoolQ \citep{clark2019boolq}, RTE \citep{wang2018glue}, HellaSwag \citep{zellers2019hellaswag}, Winogrande \citep{sakaguchi2021winogrande}, ARC Easy and Challenge \citep{clark2018think}, and OpenbookQA \citep{mihaylov2018can}. 
The results are presented in \cref{tab:performance}.

At first, we find that the perplexity effectively decreases with \BR{} and \GP{};
the value reduces across all test cases including different models, pruning methods, and datasets.
Unexpectedly, however, the perplexity rather increases when we add \CR{} despite the reduced reconstruction errors.
We also observe a similar trend in zero-shot performance for Wanda and Magnitude pruning, with mean accuracy increasing by a large margin with \BR{} and \GP{} but decreasing with \CR{}.
Interestingly, for SparseGPT, reconstruction techniques do not generally help zero-shot performance.
We hypothesize that it is because SparseGPT already conducts fairly heavy optimization compared to other methods, and applying further reconstruction on particular calibration data may not help improve zero-shot performance since it is more sensitive to distribution shift.
Furthermore, we find that such overfiting tends to occur more for LLaMA-7B than OPT-125M (see \cref{tab:testerror}).
This is possibly due to model size; \ie, given the same amount of (limited) calibration data, over-optimizing can make large models more likely to overfit and lead to poor generalization.

We can summarize our findings are as follows.
\begin{itemize}[noitemsep,nolistsep]
    \item \BR{} and \GP{} are found to be very effective in reducing perplexity in all cases; on the other hand, \CR{} often leads to overfitting, especially for large models. 
    \item This holds true for zero-shot performance as well, with only exception of SparseGPT, for which \BR{} and \GP{} do not help much in improving zero-shot performance;
    this is possibly due to the fact that SparseGPT already conducted fairly heavy optimization of remaining weights.
    It is also possible that adapting to downstream task is more prone to overfitting.
    This certainly requires more investigations.
\end{itemize}

In short, we can attempt to say without much loss of generality that ``\BR{} and \GP{} can generally help for pruning LLMs in terms of reducing perplexity''.
\section{Further Exploration}

We have seen that reconstruction techniques are useful but they can lead to undesirable overfitting.
Here we explore potential ways to alleviate this risk.
In particular, we identify that the calibration data is highly limited in two aspects:
it is too little (compared to optimization variables)\footnote{
This can be especially problematic for domain-specific LLMs, \eg, healthcare \citep{singhal2023large,luo2022biogpt} and finance \citep{wu2023bloomberggpt,yang2023fingpt}, where obtaining real-world data can be highly challenging due to privacy concerns.}
and does not represent the training data (as it is arbitrarily given);
the former is related to the general representation-generalization complexity trade-off, and the latter is about whether the reconstruction can mimic the behavior of the original model.

\begin{figure}[!t]
    \centering
    \begin{subfigure}{0.48\linewidth}
        \includegraphics[width=\linewidth]{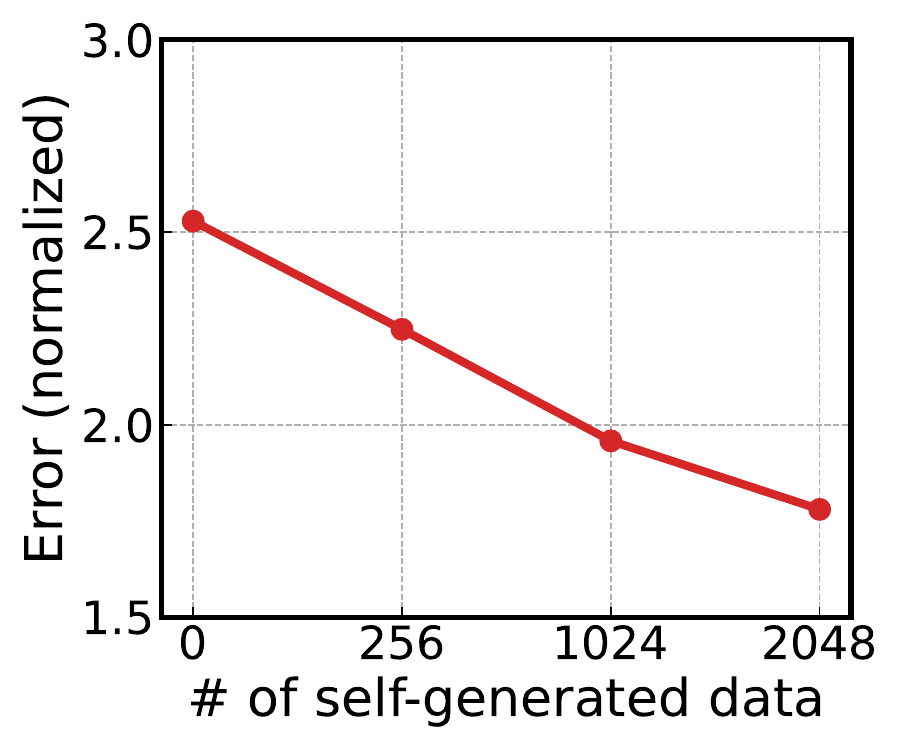}
        \caption{Test error}
         \label{fig:gendata-reconerror}
    \end{subfigure}
    \begin{subfigure}{0.48\linewidth}
        \includegraphics[width=\linewidth]{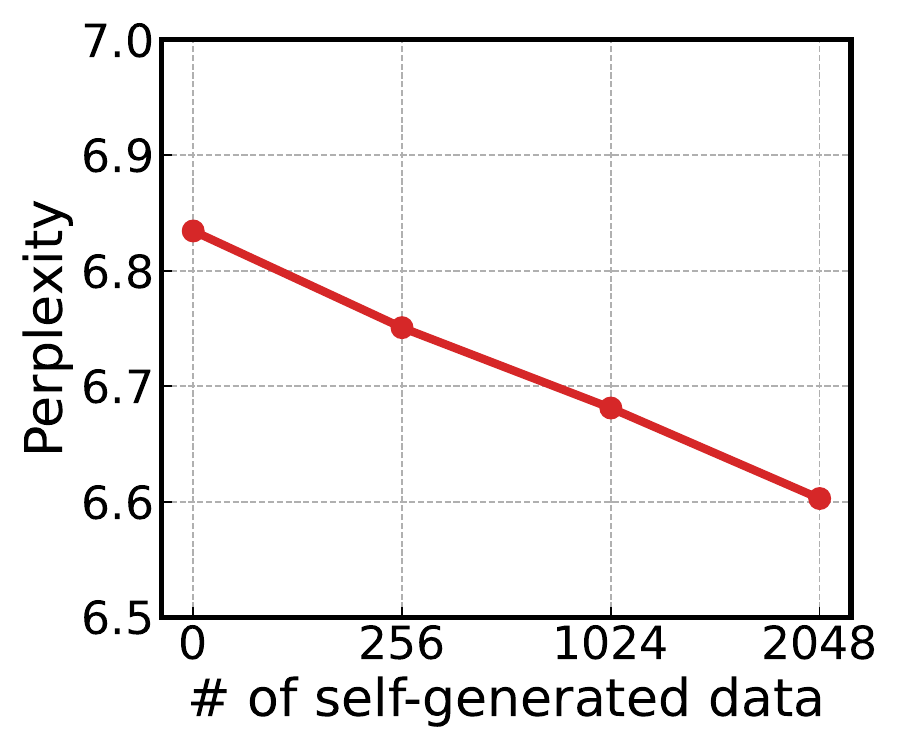}
        \caption{Perplexity}
         \label{fig:gendata-perplexity}
    \end{subfigure}
    \caption{
    Effects of self-generated calibration data on (a) reconstruction error for test data (raw-Wikitext2) and (b) perplexity for LLaMA-7B;
    they both improve with more self-generation.
    See \cref{fig:gendata-app} of \cref{sec:app-results} for more results.
    }
    \label{fig:gendata}
\end{figure}

To this end, we reflect on the fact that what we are dealing with is a \emph{generative} (language) model, meaning that we can create calibration data that is potentially much bigger in size and closer to the original distribution.
We find that this self-generation technique has recently been proposed in other contexts \citep{meng2022generating,ye2022zerogen,liu2023llm,li2024norm}, and thus, follow the process therein to produce high-quality text data.
Using that, we perform reconstruction again, and the results are reported in \cref{fig:gendata}.
We observe that making use of more self-generated calibration data (without unfairly violating the given setting) reduces both test error and perplexity, mitigating overfitting quite effectively.

\section{Conclusion}

In this work, we take a close look at the current practice of minimizing reconstruction errors for pruning LLMs.
We first find that with various reconstruction techniques, one can reduce the error quite significantly and improve quality of pruning results on both language perplexity and zero-shot accuracy.
Nevertheless, it turns out that decreasing error as it is now is not always desirable since it may cause overfitting calibration data.
We present initial results that this issue can be potentially mitigated by self-generating calibration data.
There are many remaining possibilities, and we believe our findings suggest opportunities for future work.
\section{Limitations}

There remain several limitations in our experiments and we plan to address these in future work.
First, our main experiments are limited to LLaMA-7B and OPT-125M.
We intend to scale up our experiments to much larger models of up to 70B parameters and different architectures including Mixtral \citep{jiang2024mixtral} or Gemma \citep{team2024gemma}.
Next, reconstruction techniques \BR{}, \GP{}, and \CR{} require additional memory compared to \LR{}, although they still use much less memory compared to model-level reconstruction of solving (\ref{eq:reconstruction}) (see \cref{sec:app-results} for the details).
We plan to introduce parameter-efficient optimization \citep{hu2021lora} to alleviate this increased memory burden.

Although the self-generation of calibration data effectively mitigates overfitting, it requires more computation for reconstruction.
Finally, we find that some portions of the generated texts are far from plain English texts and thus may not serve as good calibration data (see \cref{tab:gendata-example} of \cref{app:gendata} for the examples).
In this regard, we believe that reducing the number of these irrelevant examples and generating only a few number of high-quality texts can be a potential way to improve performance and increase efficiency.
\section*{Acknowledgements}
This work was partly supported by the Institute of Information \& communications Technology Planning \& Evaluation (IITP) grant funded by the Korean government (MSIT) (RS-2019-II191906, Artificial Intelligence Graduate School Program (POSTECH); RS-2022-II220959/No.2022-0-00959, (part2) Few-Shot learning of Causal Inference in Vision and Language for Decision Making; RS-2024-00338140, Development of Learning and Utilization Technology to Reflect Sustainability of Generative Language Models and Up-to-Dateness over Time) and the National Research Foundation of Korea (NRF) grant funded by the Korean government (MSIT) (RS-2023-00210466, RS-2023-00265444, RS2023-0021371).
Sungbin Shin was supported by Kwanjeong Educational Foundation Scholarship.

\bibliography{custom}

\appendix

\section{Experimental Details}
\label{sec:app-details}

\paragraph{Experiment configurations}
We run our experiments with a single A100 GPU having $80$GB of memory.
For \BR{} and \CR{}, we run the Adam optimizer for $10$ epochs with a batch size of $8$, without weight decay or gradient clipping.
The learning rate is set to $0.0002$ and decays linearly following \citet{guo2024ebft}.
For evaluating the performance on downstream tasks, we use the EleutherAI-evalharness framework \citep{eval-harness}.

\paragraph{Calculation of normalized reconstruction error}
The reconstruction error for $i$-th block is calculated as $\frac{1}{N H T} \| g_i (\bar{w}_i; \bar{x}_i) - g_i(m_i \odot w_i; x_i)  \|_2^2 $ where $N, H, T$ each represent the number of calibration data, hidden dimension, and the token length. $\bar{x}_i, x_i$ represent the inputs coming from dense and sparse blocks respectively.

\paragraph{Licenses and uses of models and datasets}
LLaMA \citep{touvron2023llama} and OPT \citep{zhang2022opt} are released under non-commercial bespoke 
licenses.
raw-Wikitext2 \citep{merity2016pointer}, PTB \citep{marcus1994penn}, and C4 \citep{raffel2020exploring} are released under CC BY-SA 4.0, LDC user agreement, and ODC-By.
BoolQ \citep{clark2019boolq}, RTE \citep{wang2018glue}, HellaSwag \citep{zellers2019hellaswag}, Winogrande \citep{sakaguchi2021winogrande}, ARC \citep{clark2018think}, and OpeenbookQA \citep{mihaylov2018can} are released under CC BY-SA 3.0, Apache 2.0, MIT License, Apache 2.0, CC BY-SA 4.0, and Apache 2.0 respectively.
We confirm that these models and datasets are used for their intended use and the data does not contain personal information.
EleutherAI-evalharness framework is released under the MIT License.

\section{Additional Results}
\label{sec:app-results}

\begin{figure}[!th]
    \centering
    \begin{subfigure}{0.32\linewidth}
        \includegraphics[width=\linewidth]{figures/reconerror/llama_sparsegpt_all.pdf}
        \caption{SparseGPT}
         \label{fig:recon-error-app-sparsegpt}
    \end{subfigure}
    \begin{subfigure}{0.32\linewidth}
        \includegraphics[width=\linewidth]{figures/reconerror/llama_wanda_all.pdf}
        \caption{Wanda}
         \label{fig:recon-error-app-wanda}
    \end{subfigure}
    \begin{subfigure}{0.32\linewidth}
        \includegraphics[width=\linewidth]{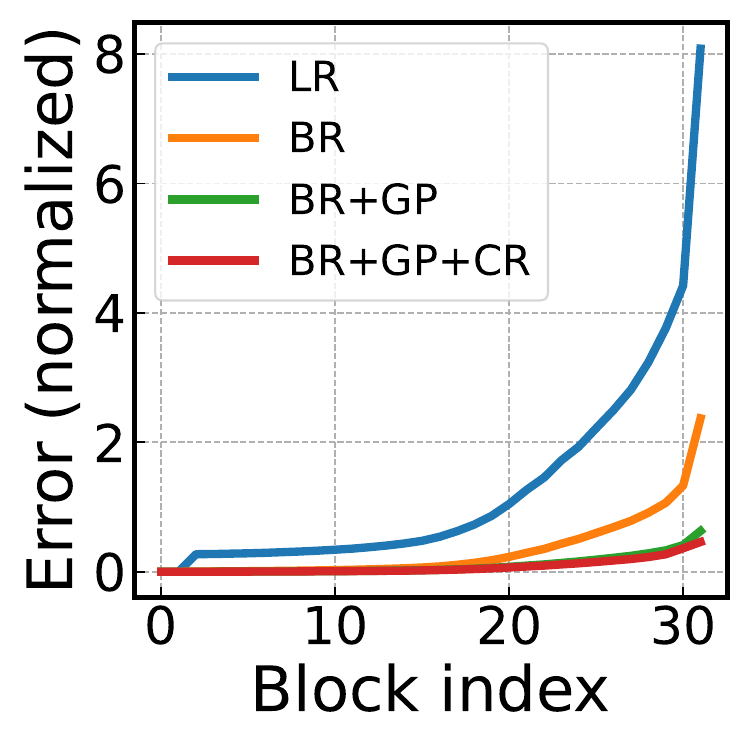}
        \caption{Magnitude}
         \label{fig:recon-error-app-mag}
    \end{subfigure}
    \caption{
    Results of reconstruction techniques for LLaMA-7B.
    They constantly reduce the compounding errors, achieving a significant decrease at the final block ($87\% \sim 94\%$).
    }
    \label{fig:recon-error-app}
\end{figure}

\begin{figure}[!th]
    \centering
    \begin{subfigure}{0.32\linewidth}
        \includegraphics[width=\linewidth]{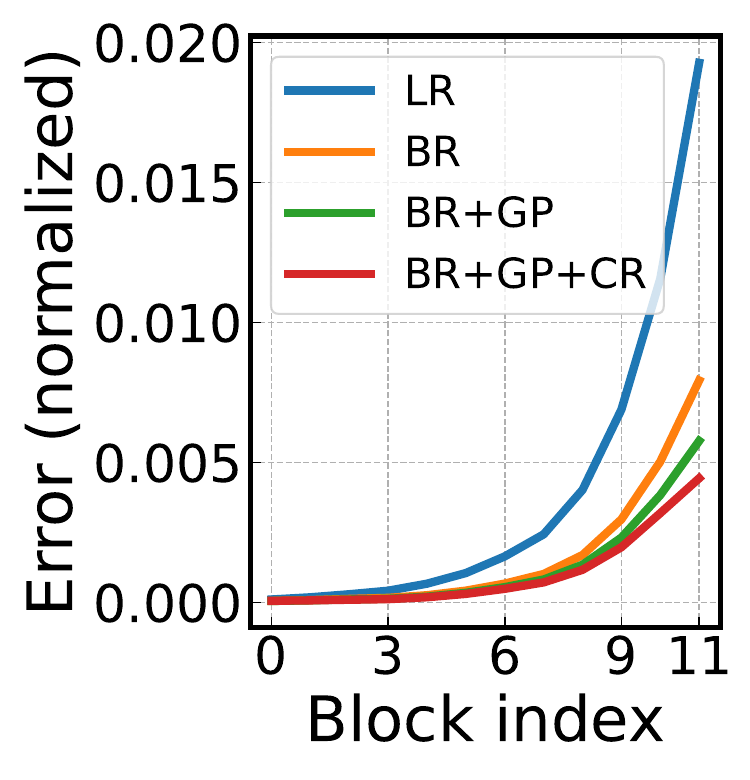}
        \caption{SparseGPT}
         \label{fig:recon-error-app-sparsegpt-opt}
    \end{subfigure}
    \begin{subfigure}{0.32\linewidth}
        \includegraphics[width=\linewidth]{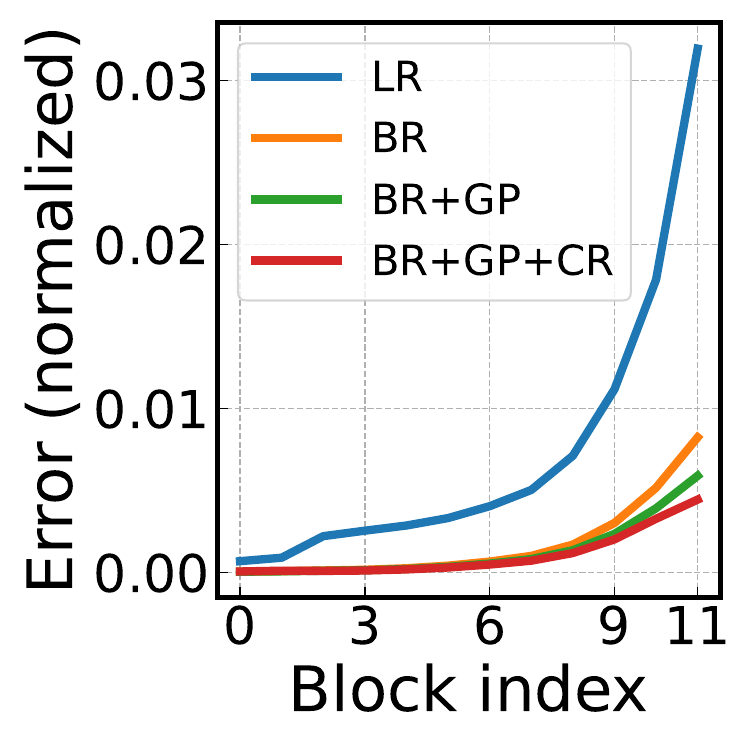}
        \caption{Wanda}
         \label{fig:recon-error-app-wanda-opt}
    \end{subfigure}
    \begin{subfigure}{0.32\linewidth}
        \includegraphics[width=\linewidth]{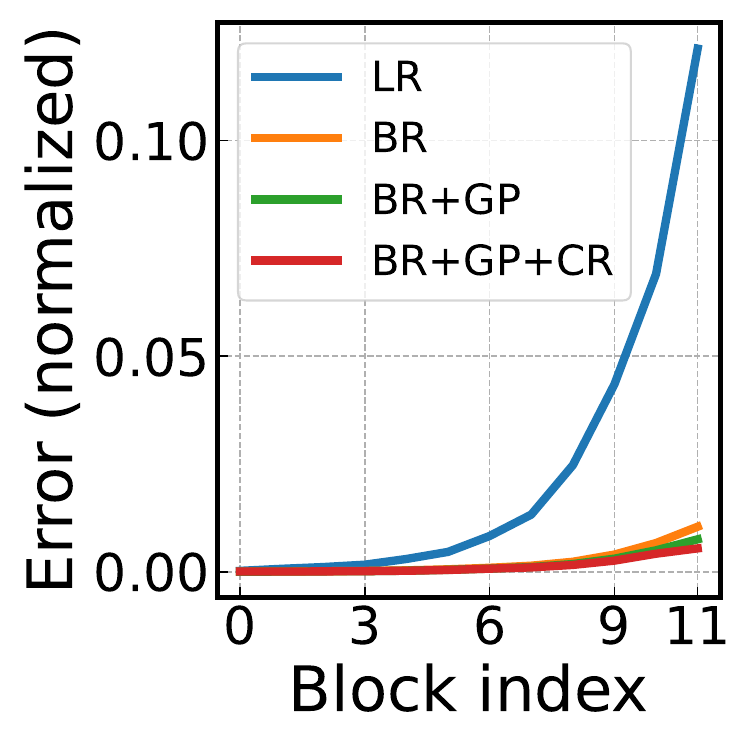}
        \caption{Magnitude}
         \label{fig:recon-error-app-mag-opt}
    \end{subfigure}
    \caption{
    Results of reconstruction techniques for OPT-125M.
    They constantly reduce the compounding errors, achieving a significant decrease at the final block ($79\% \sim 96\%$).
    }
    \label{fig:recon-error-app-opt}
\end{figure}

\begin{table*}[!t]
    \centering
    \resizebox{\linewidth}{!}{
    \begin{tabular}{c c | c | c c c c| c c c c c c c c}
      \toprule
      \multirow{2}{*}{Pruner} & \multirow{2}{*}{Reconstruction} & \multirow{2}{*}{Error (normalized)} & \multicolumn{4}{c|}{Perplexity} & \multicolumn{8}{c}{Zero-shot accuracy} \\
      & & & Wiki & PTB & C4 & Mean & BoolQ & RTE & HellaSwag & WinoGrande & ARC-e & ARC-c & OpenbookQA & Mean \\
      \midrule
      Dense & $-$ & $-$ & $27.66$ & $38.99$ & $26.56$ & $31.07$ &  $55.44$ & $50.18$ & $29.19$ & $50.20$ & $43.60$ & $19.03$ & $16.6$ & $37.75$\\
      \midrule
      \multirow{4}{*}{SparseGPT}& \LR{} & $0.019$ & $36.35$ & $54.93$ & $33.12$ & $41.47$ & $\underline{61.31}$ & $\underline{48.01}$ & $28.29$ & $\underline{53.28}$ & $40.19$ & $19.28$ & $15.60$ & $\mathbf{38.00}$\\
& \BR{} & $0.008$ & $31.94$ & $45.75$ & $29.91$ & $35.87$ & $60.49$ & $47.65$ & $28.44$ & $51.38$ & $42.17$ & $\underline{19.88}$ & $14.60$ & $37.80$\\
& \BR{}+\GP{} & $0.006$ & $31.57$ & $45.52$ & $29.81$ & $35.63$ & $60.18$ & $45.13$ & $28.53$ & $52.17$ & $\underline{42.63}$ & $19.62$ & $14.80$ & $37.58$\\
& \BR{}+\GP{}+\CR{} & $\mathbf{0.004}$ & $\underline{30.86}$ & $\underline{44.61}$ & $\underline{29.45}$ & $\mathbf{34.97}$ & $60.31$ & $46.21$ & $\underline{28.64}$ & $51.07$ & $42.63$ & $19.71$ & $\underline{15.80}$ & $37.77$\\
      \midrule
      \multirow{4}{*}{Wanda}& \LR{} & $0.032$ & $39.00$ & $56.27$ & $34.62$ & $43.30$ & $\underline{62.05}$ & $\underline{48.38}$ & $28.31$ & $\underline{52.01}$ & $39.56$ & $\underline{19.62}$ & $14.20$ & $37.73$\\
& \BR{} & $0.008$ & $31.55$ & $46.17$ & $29.89$ & $35.87$ & $60.24$ & $47.65$ & $28.34$ & $50.20$ & $41.50$ & $19.54$ & $15.00$ & $37.50$\\
& \BR{}+\GP{} & $0.006$ & $31.18$ & $45.47$ & $29.67$ & $35.44$ & $59.85$ & $48.01$ & $28.66$ & $51.54$ & $41.71$ & $19.28$ & $\underline{16.20}$ & $\mathbf{37.89}$\\
& \BR{}+\GP{}+\CR{} & $\mathbf{0.004}$ & $\underline{30.59}$ & $\underline{44.80}$ & $\underline{29.33}$ & $\mathbf{34.91}$ & $58.81$ & $45.85$ & $\underline{28.68}$ & $50.99$ & $\underline{42.34}$ & $19.03$ & $15.00$ & $37.24$\\
      \midrule
      \multirow{4}{*}{Magnitude} & \LR{} & $0.121$ & $193.36$ & $276.15$ & $141.01$ & $203.5$ & $\underline{60.55}$ & $\underline{53.43}$ & $27.32$ & $\underline{52.57}$ & $33.04$ & $\underline{19.97}$ & $14.20$ & $37.30$\\
& \BR{} & $0.010$ & $36.06$ & $49.15$ & $31.63$ & $38.95$ & $58.99$ & $48.38$ & $28.35$ & $51.22$ & $41.20$ & $19.88$ & $\underline{15.80}$ & $37.69$\\
& \BR{}+\GP{} & $0.008$ & $35.56$ & $48.17$ & $31.75$ & $38.50$ & $58.20$ & $49.46$ & $28.44$ & $51.54$ & $\underline{42.26}$ & $19.88$ & $15.20$ & $\mathbf{37.85}$\\
& \BR{}+\GP{}+\CR{} & $\mathbf{0.005}$ & $\underline{33.76}$ & $\underline{46.84}$ & $\underline{30.88}$ & $\mathbf{37.16}$ & $57.28$ & $45.49$ & $\underline{28.53}$ & $51.93$ & $42.00$ & $19.97$ & $15.60$ & $37.26$\\
      \bottomrule
  \end{tabular}
  }
  \caption{
  Effects of different reconstruction techniques on error, perplexity, and zero-shot accuracy for OPT-125M.
  \textbf{Bold} and \underline{underline} refer to best in general and task-specific.
  }
  \label{tab:performance-opt}
\end{table*}

\begin{table*}
   \centering
   \captionsetup[subtable]{position = below}
   \begin{subtable}{0.48\linewidth}
       \centering
       \resizebox{\linewidth}{!}{
    \begin{tabular}{c c c c c c}
      \toprule
     \multirow{2}{*}{Pruner} & \multirow{2}{*}{\CR{}} & \multicolumn{4}{c}{Error (normalized)} \\
      & & Calib & Test (Wiki) & Test (PTB) & Tets (C4) \\
      \midrule
      \multirow{2}{*}{SparseGPT} & X & $0.006$ & $0.0083$ & $0.009$ & $0.0065$\\
& O & $\mathbf{0.004}$ & $\mathbf{0.0078}$ & $\mathbf{0.0083}$ & $\mathbf{0.0061}$\\
      \midrule
      \multirow{2}{*}{Wanda} 
& X & $0.006$ & $0.008$ & $0.0088$ & $0.0061$\\
& O & $\mathbf{0.004}$ & $\mathbf{0.0076}$ & $\mathbf{0.0082}$ & $\mathbf{0.0058}$\\
      \midrule
      \multirow{2}{*}{Magnitude} 
& X & $0.008$ & $0.0109$ & $0.0115$ & $0.0125$\\
& O & $\mathbf{0.005}$ & $\mathbf{0.0102}$ & $\mathbf{0.0111}$ & $\mathbf{0.0099}$\\
      \bottomrule
  \end{tabular}
  }
   \caption{OPT-125M}
   \label{tab:testerror-opt-app}
   \end{subtable}%
   \begin{subtable}{0.48\linewidth}
       \centering
       \resizebox{\linewidth}{!}{
    \begin{tabular}{c c c c c c}
      \toprule
      \multirow{2}{*}{Pruner} & \multirow{2}{*}{\CR{}} & \multicolumn{4}{c}{Error (normalized)} \\
      & & Calib & Test (Wiki) & Test (PTB) & Tets (C4) \\
      \midrule
      \multirow{2}{*}{SparseGPT} & X & $0.48$ & $\mathbf{2.30}$ & $\mathbf{2.29}$ & $\mathbf{1.99}$\\
& O & $\mathbf{0.37}$ & $2.53$ & $2.60$ & $2.31$\\
      \midrule
      \multirow{2}{*}{Wanda} 
& X & $0.51$ & $\mathbf{2.23}$ & $\mathbf{2.29}$ & $\mathbf{1.98}$\\
& O & $\mathbf{0.38}$ & $2.48$ & $2.86$ & $2.31$\\
      \midrule
      \multirow{2}{*}{Magnitude} 
& X & $0.63$ & $\mathbf{2.42}$ & $\mathbf{2.72}$ & $\mathbf{2.21}$\\
& O & $\mathbf{0.46}$ & $2.55$ & $3.03$ & $2.40$\\
      \bottomrule
  \end{tabular}
  }
   \caption{LLaMA-7B}
   \label{tab:testerror-llama7B-app}
   \end{subtable}%
   \caption{
   Reconstruction errors of OPT-125M and LLaMA-7B on test data (raw-Wikitext2) as well as calibration data.
   Overfitting by \CR{} is only observed for the larger LLaMA-7B model.
   }
   \label{tab:testerror-app}
\end{table*}

\begin{figure*}[!t]
    \centering
    \begin{subfigure}{0.32\linewidth}
        \includegraphics[width=0.48\linewidth]{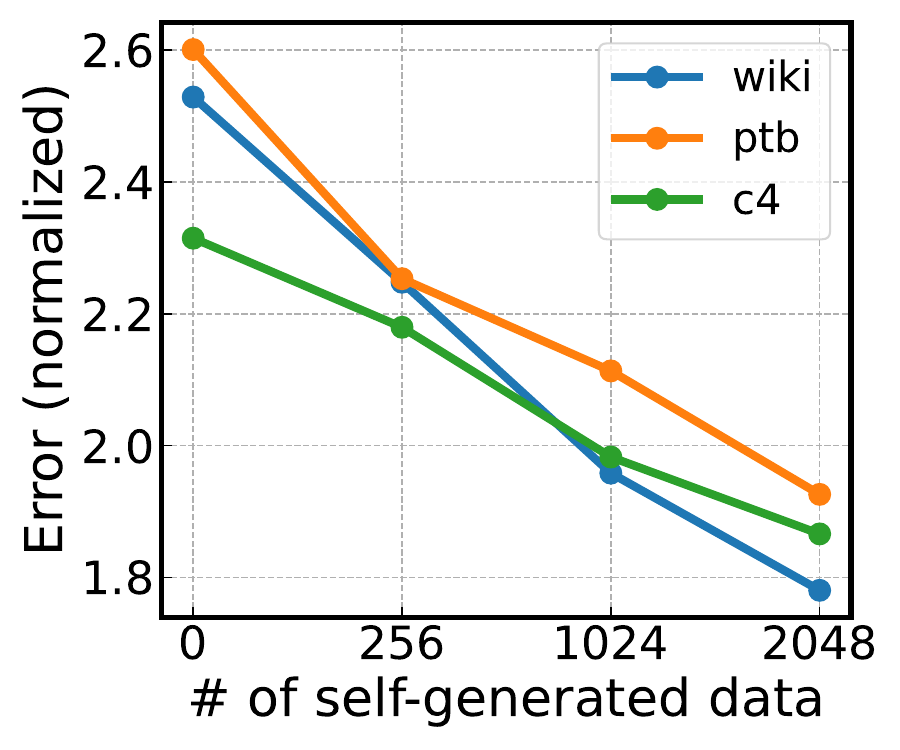}
        \includegraphics[width=0.48\linewidth]{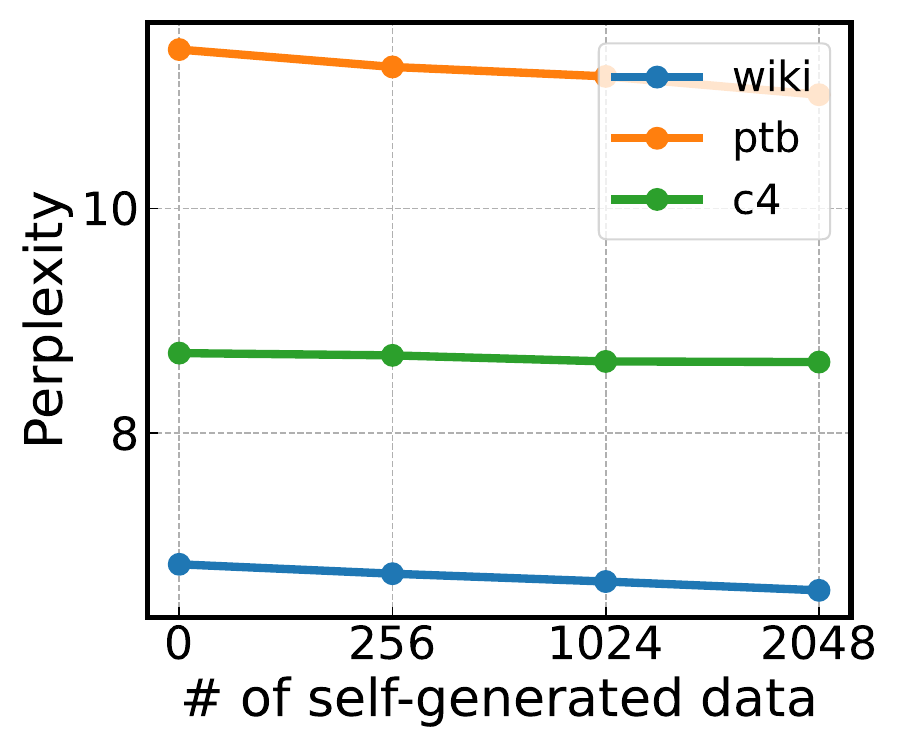}
        \caption{SparseGPT}
         \label{fig:gendata-sgpt}
    \end{subfigure}
    \begin{subfigure}{0.32\linewidth}
        \includegraphics[width=0.48\linewidth]{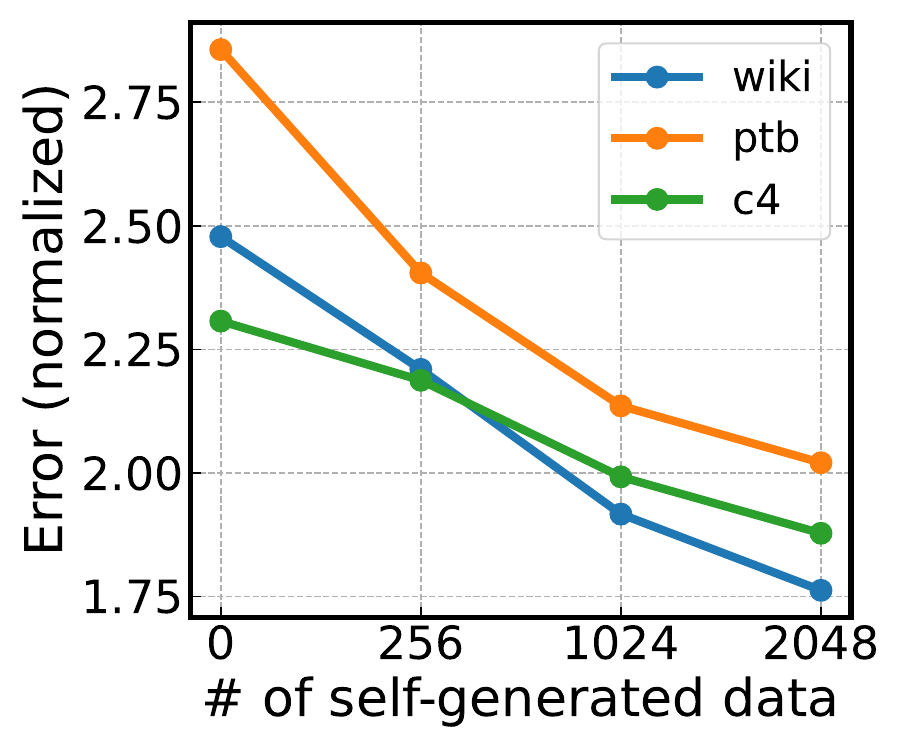}
        \includegraphics[width=0.48\linewidth]{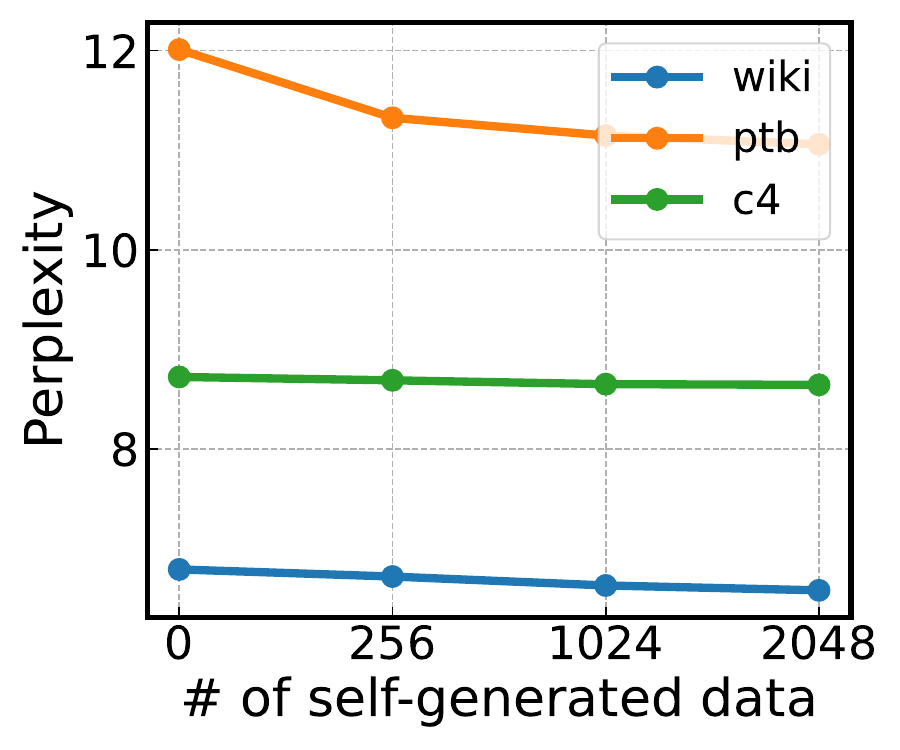}
        \caption{Wanda}
         \label{fig:gendata-wanda}
    \end{subfigure}
    \begin{subfigure}{0.32\linewidth}
        \includegraphics[width=0.48\linewidth]{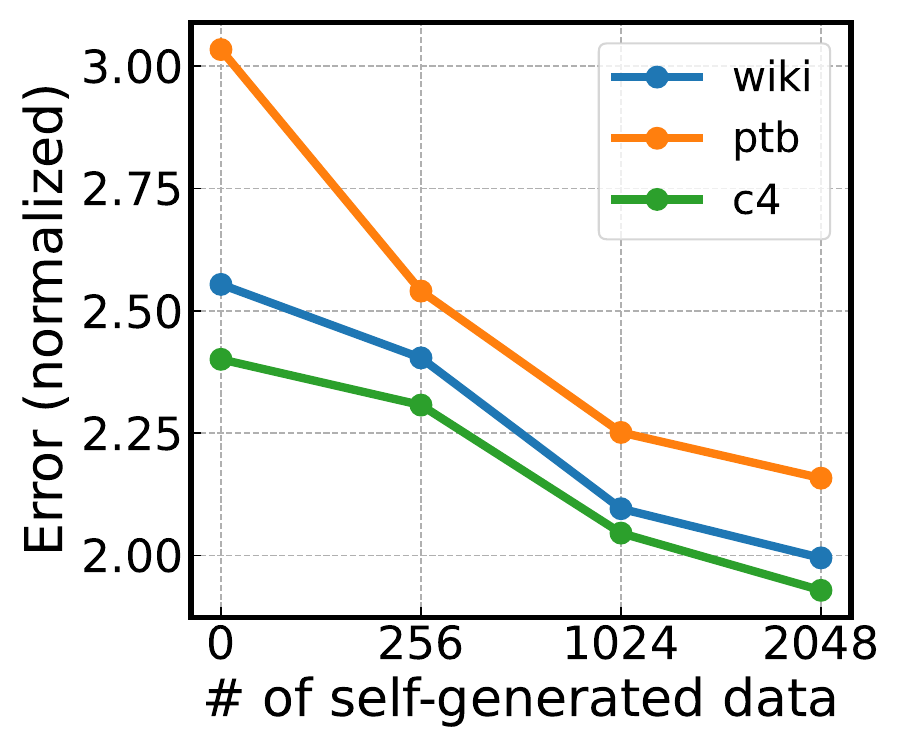}
        \includegraphics[width=0.48\linewidth]{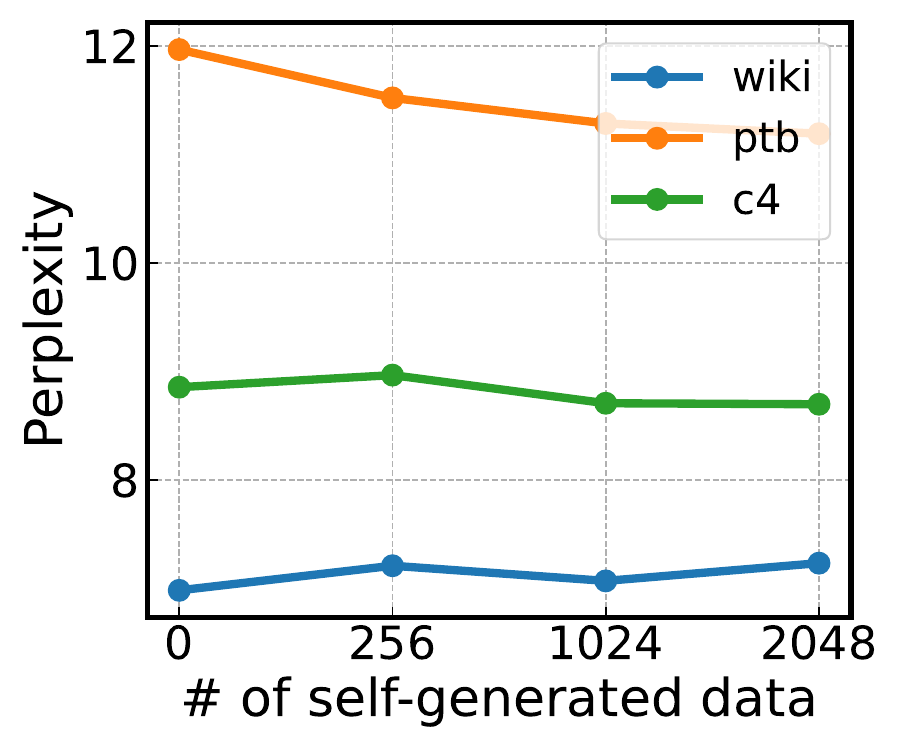}
        \caption{Magnitude}
         \label{fig:gendata-mag}
    \end{subfigure}
    \caption{
     Effects of self-generated calibration data on reconstruction error for test data and perplexity for LLaMA-7B;
     they both improve with more self-generation.
    }
    \label{fig:gendata-app}
\end{figure*}

\begin{table*}[!t]
    \centering
    \resizebox{\linewidth}{!}{
    \begin{tabular}{c c}
      \toprule
      Example number & Text \\
      \midrule
      1 & \emph{Americas, and the U.K., while 18 other countries have legalized the medical use of cannabis. The latest announcement is a win for Canadians ...} \\
      2 & \emph{apprehension of the inevitability of death? And, therefore, how could such a person come to believe ...} \\
      3 & \emph{`\#' + this.currentID + .\"'\textbackslash n \};\textbackslash n\textbackslash n return \{\textbackslash n next: next,\textbackslash n previous: previous,\textbackslash n\}...} \\
      4 & \emph{Picker.setSelected(false);\textbackslash n \textbackslash n actionPhrasesTableModel.fireTableDataChanged();\textbackslash n} ... \\
      \bottomrule
  \end{tabular}
  }
  \caption{
  Examples of self-generated data.
  }
  \label{tab:gendata-example}
\end{table*}

\begin{table*}
   \centering
       \centering
       \resizebox{0.6\linewidth}{!}{
    \begin{tabular}{c c c c c c}
      \toprule
      & \LR{} & \BR{} & \BR{}+\GP{} & \BR{}+\GP{}+\CR{} & Full fine-tuning
      \\
      \midrule
      peak memory (GB) & $3.9$ & $5.7$ & $5.7$ & $10.6$ & $>100$ \\
      \bottomrule
  \end{tabular}
  }
   \caption{
   Peak GPU memory for LLaMA-7B and sparseGPT.
   Compared to \LR{}, reconstruction techniques incur additional GPU memory but it is quite marginal compared to fine-tuning the full model.
   The results are obtained with the batch size of $8$ and gradient accumulation.
   For full fine-tuning, the results are from \citet{malladi2023fine}.
   }
   \label{tab:recon-memory}
\end{table*}

\begin{table*}
   \centering
       \centering
       \resizebox{0.6\linewidth}{!}{
    \begin{tabular}{c c c c c c}
      \toprule
      Pruning block & \LR{} & \BR{} & \BR{}+\GP{} & \BR{}+\GP{}+\CR{}
      \\
      \midrule
      Attention & $32.82$ & $30.15$ & $29.97$ & $29.64$ \\
      Feed-forward & $30.69$ & $29.23$ & $28.89$ & $28.73$ \\
      All & $36.35$ & $31.94$ & $31.57$ & $30.86$ \\
      \bottomrule
  \end{tabular}
  }
   \caption{
   Effects of pruning block for different reconstruction techniques. 
   Here, we prune either attention or feed-forward block to $50\%$ sparsity and measure the perplexity on raw-Wikitext2.
   Pruning only the attention block leads to worse performance compared to pruning only the feed-forward block.
   The results are for OPT-125m with sparseGPT.
   }
   \label{tab:attn-vs-feedforward}
\end{table*}

\paragraph{More results on the reconstruction techniques}

Effects of reconstruction techniques on reducing the error for LLaMA-7B and OPT-125M are presented in \cref{fig:recon-error-app,fig:recon-error-app-opt} respectively.
It is clearly observed that different reconstruction techniques significantly reduce the error for all cases.

Effects of reconstruction techniques on performance for OPT-125M are presented in \cref{tab:performance-opt}.
Different techniques effectively improve the performance on perplexity and downstream tasks, with the exception of overfitting for \CR{} on downstream tasks.

\paragraph{More results on self-generated data}

Reconstruction error on calibration data and test data for OPT-125M and LLaMA-7B are presented in \cref{tab:testerror-app}.
Decreased error for calibration data leads to decreased error for test data for OPT-125M, but leads to increased test error for LLaMA-7B.

Effects of self-generated calibration data are presented in \cref{fig:gendata-app}.
In most cases, more number of self-generated data leads to decreased test error and perplexity.

\paragraph{Memory consumption of reconstruction techniques}

Solving (\ref{eq:reconstruction}) directly can be memory-intensive, thus many recent work suggest divide-and-conquer such as \LR{} and \BR{}.
In the work of \citet{frantar2023sparsegpt}, the authors show that for the 175B parameter OPT model it requires at least five A100 GPUs of 80GB, whereas by using \LR{} it reduces down to a single A100 GPU of 80GB.
In our experiments, for Llama-7B, both \LR{} and \BR{}+\GP{}+\CR{} can all be done on a commodity 3090 GPU of 24GB memory;
it requires more than 100GB to perform full fine-tuning of LLaMA-7B \citep{malladi2023fine}.
In theory, optimizing more parameters can incur more memory footprints, and thus, in the order of \LR{} $=$ \GP{} $<$ \BR{} $<$ \CR{}, there will be more memory usage.

The exact amount depends on the specific model.
To provide solid evidence, we ran profiling peak GPU memory for LLaMA-7B with the batch size of $8$ (see \cref{tab:recon-memory} for the results).
Compared to \LR{}, reconstruction techniques surely incur additional GPU memory, however, (i) it is quite marginal compared to fine-tuning the full model, and (ii) it could be reduced further by introducing memory reduction techniques in practice such as CPU offloading and gradient checkpointing.

\paragraph{Pruning attention vs. feed-forward}

We also investigated the effects of only pruning attention vs. feed-forward blocks for different reconstruction techniques.
Here, we conducted experiments for OPT-125m and SparseGPT by pruning either attention or feed-forward blocks to $50\%$ sparsity and measuring the perplexity on raw-Wikitext2.
The results are provided in \cref{tab:attn-vs-feedforward}.
We first observe that pruning both attention and feed-forward yields the largest performance drop.
Also, we find that pruning only the attention block leads to worse performance compared to pruning only the feed-forward block, which is consistent with the findings in the previous work \citep{namburi2023cost}.
Interestingly, we find that reconstruction techniques can be more effective for cases with poor performance; \ie, in the order of pruning all blocks > pruning attention > pruning feed-forward, \BR{}, \GP{}, \CR{} reconstruction techniques yield more reduction in perplexity (which is good by itself).

\section{Details on Self-generation of Calibration Data}

\label{app:gendata}

We generate additional calibration data from the original dense model.
Here, we sample $10240$ number of English texts each containing $2048$ tokens.
Specifically, we first randomly choose the initial token and generate four subsequent tokens by deterministically selecting top-$1$ predictions, similar to \citet{liu2023llm}.
Here, we resample the tokens if the generated texts are not detected as English.
Then, we stochastically generate the remaining tokens until the <EOS> token is produced or the sequence length exceeds $2048$.
Finally, the additional calibration data can be obtained by sampling a subset of generated texts and randomly selecting the intermediate $1024$ tokens for each text.

Examples of self-generated texts are presented in \cref{tab:gendata-example}.
Examples $1$ and $2$ are plain English texts and can serve as good calibration data.
However, we observe that programming codes such as examples $3$ and $4$ are often generated, which might not serve as good calibration data for improving the perplexity for English texts or accuracy for downstream tasks which are not related to code generation.
In this regard, we believe that generating only a few number of high-quality texts can lead to improved performance while reducing computational costs.

Here, the generated data do not contain personal information or offensive content.

\end{document}